\def\year{2018}\relax
\documentclass[letterpaper]{article} 
\usepackage{aaai18}  
\usepackage{times}  
\usepackage{helvet}  
\usepackage{courier}  
\usepackage{url}  
\usepackage{graphicx}  
\usepackage{multirow}
\usepackage{longtable}
\usepackage{supertabular,booktabs}
\usepackage{color, colortbl}
\usepackage{tikz}
\usepackage{amsmath}
\usetikzlibrary{matrix}
\usetikzlibrary{arrows, decorations.markings}
\definecolor{Gray}{gray}{0.9}

\begin{document}
\title{Computation Error Analysis of Block Floating Point Arithmetic Oriented Convolution Neural Network Accelerator Design}
\author{ Zhourui Song$^1$,   Zhenyu Liu$^2$ and Dongsheng Wang$^2$ \\
$^1$ School of Cyberspace Security, Beijing University of Posts and Telecommunications, Beijing, 100876, China \\
$^2$ RIIT, Tsinghua University, Beijing, 100084, China \\
Email: songzr@bupt.edu.cn liuzhenyu73@mail.tsinghua.edu.cn wds@tsinghua.edu.cn\\
}
\maketitle
\begin{abstract}
The heavy burdens of computation and off-chip traffic impede deploying the large scale convolution neural network on embedded platforms. As CNN is attributed to the strong endurance to computation errors, employing block floating point (BFP) arithmetics in CNN accelerators could save the hardware cost and data traffics efficiently, while maintaining the classification accuracy. In this paper, we verify the effects of word width definitions in BFP to the CNN performance without retraining. Several typical CNN models, including VGG16, ResNet-18, ResNet-50 and GoogLeNet, were tested in this paper. Experiments revealed that 8-bit mantissa, including sign bit, in BFP representation merely induced less than 0.3\% accuracy loss. In addition, we investigate the computational errors in theory and develop the noise-to-signal ratio (NSR) upper bound, which provides the promising guidance for BFP based CNN engine design.
\end{abstract}
\section {Introduction}
\noindent Convolutional neural networks (CNNs) have achieved state-of-art performance in many artificial intelligence tasks, especially in image recognition \cite{ciregan2012multi}  \cite{russakovsky2015imagenet}, nature language processing\cite{kim2014convolutional}\cite{goldberg2016primer}, strategic planning\cite{silver2016mastering}, etc. This success is partially facilitated by the advance of computation infrastructure. With GPU clusters, large-scale CNNs can be deployed eventhough they are attributed as memory-and-computation-intensive and resource-consuming\cite{li2016high}. However, when deploying CNNs in data center, GPU clusters is not the first preference because of the low power efficiency of GPU. Therefore, promoting energy efficiency became one prominent target in CNN accelerator design. Researchers  have been committed to exploring how to deploy CNNs on FPGAs \cite{ovtcharov2015accelerating} , or designing AISCs\cite{jouppi2017datacenter}, as they prossesses higher energy efficiency due to their specific architecture.

To transplant CNNs on FPGA, two serious issues, i.e., off-chip traffic bottleneck and huge amount of floating-point arithmetics overhead, need to be addressed. The off-chip traffic stems from that, for large scale networks, the  feature maps and the network parameters must be stored in the off-chip SDRAM. The frequent accesses to these datum induces no-trivial bandwidth requirements.
Secondly, as the hardwired floating-point modules are not always equipped in FPGA, employing the floating point operations in FPGA CNN accelerator degrades both of the throughput and the energy efficiency severely.

In this paper, we proposed a block floating point (BFP) based convolution implementation. BFP representation can be attributed as a special case of floating point representation where numbers within a block share a common exponent. Hence, BFP possesses the high dynamic range of floating point representation as it has exponent part. On the other hand, the computation complexity of two BFP blocks is reduced to the degree of integer representation. Experiments in our paper revealed that even with 7-bit mantissa, we can sacrifice less than 0.3\% accuracy loss when implementing the large scale network models, such as  VGG-16\cite{simonyan2014very}, ResNet-18, ResNet-50\cite{he2016deep} and GoogLeNet\cite{Szegedy_2015_CVPR}. It should be noted that, no retraining is required in our experiments. That is the original models can be deployed in our BFP based accelerator directly. 
Finally, this paper proposed an analytical model to derive the NSR upper bound of the finite word-length BFP arithmetics, which supports  the verification of hardwired CNN accelerator design efficiently .

The rest of this paper is organized as follow. Section 2 presented related works. Section 3 discussed the details of applying BFP in CNNs. Section 4 expounded the theoretical NSR model of BFP. Section 5 verified the performance of BFP oriented CNN on GoogLeNet, VGG-16, ResNet-18, ResNet-50, mnist and cifar10. We also illustrated the efficiency of our NSR model by using VGG-16 network. Section 6 summarized the whole paper.

\section{Related Works}
Methods like data reusing, compression and trimming have been developed to meet the bandwidth requirement of FPGA. \cite{chen2017eyeriss} \cite{karam2017memory} proposed a row stationary data flow on 168 processing elements that improve the energy efficiency by reusing data locally. \cite{zhang2015optimizing} develop the roofline model to analyze the computation and memory requirements of a specified CNN model, and then to identify the optimal solution on the provided FPGA platform. Sparsifying CNN model's parameters is another popular solution. \cite{han2015deep} proposed a three-stage compression method, namely pruning, trained quantization and Huffman coding that significantly reduced the size of DNNs without decrease in accuracy. However, retraining of the sparse model is time consuming, and the entropy decoding of model parameter causes additional  delay when accessing these parameters. That is, the CNN accelerator's throughput is degraded. Trimming \cite{han2015learning}\cite{parashar2017scnn} also suffers from retraining as it is required to find the important connections and abandon the others.

Researchers have committed to replacing 32-bit floating point number format with fixed point number format. \cite{page2016fpga} utilized singular value decomposition on dense layers and limited precision fixed-point representations of convolutional weights, at the cost of less than 0.2\% decrease in accuracy on MNIST, CIFAR-10 and SVHN with 3-bit integer and 6-bit fraction fixed point format. Rounding model has also drawn attention. \cite{gupta2015deep} proposed that deep networks can be trained in 16-bit fixed point representation with stochastic rounding. However, the common weakness of  the above methods is that they all require retrain to amended parameters, while retrain is very expensive. In addition, when applied in deep neural networks, the quick growth of word width requirement consumes more chip area, power, and bandwidth, which becomes the hindrance of employing integer arithmetic in complex network models. For example, \cite{hill2016rethinking} proved that GoogLeNet acquires 40-bit fixed point representation to maintain an acceptable accuracy using stochastic rounding.

\cite{mellempudi2017mixed} proposed a method that divide weights and activations into clusters, and each cluster holds a joint scaling factor. Therefore, the numbers in the same cluster can be represented by a integer index. The subsequent convolution operation can be carried out in the integer domain. They designed a system that utilizes 8-bit integer achieving 6\% decrease in ResNet-101 top-1 accuracy without retraining. This scheme partly eliminated the floating point operations. In specific, the scaling procedure is still carried out with floating point arithmetics, which even include the divide and the root operations.

\section{Block-Floating Point Arithmetic \\Oriented CNN}
\subsection{Definition of Block Floating Point Arithmetic }
\noindent With block floating point representation, n numbers belonging to a block share the common scaling factor, i.e., the block exponent. The block exponent is determined by the largest magnitude in the block, and smaller ones will be right shifted to align, which is called block formatting.

At first, we provide the associated nomenclature to clarify our statement. For a cluster of numbers, denoted as $\mathbf X$, $x_{i}$ is the $i$th element of $\mathbf X$, $m_i$ and $e_i$ are the mantissa and exponent part of $x_{i}$. When $\mathbf X$ is block formatted into $\mathbf X'$, the mantissa part and block exponent is written as $\mathbf M'_{\mathbf X}$ and $ \varepsilon_{\mathbf X}$, respectively.

For example, given a block $\mathbf X$ that contains $N$ floating numbers, $\mathbf X$ can be expressed as
\begin {eqnarray}
\mathbf X & = &  \begin{pmatrix} x_1,& \cdots & x_i, & \cdots & x_N \end{pmatrix} \nonumber \\
& = & \begin{pmatrix} m_1\times 2^{e_1}, & \cdots ,&  m_i\times 2^{e_i}, & \cdots ,&  m_N\times 2^{e_N}\end{pmatrix} \nonumber
\end{eqnarray}
With BFP representation, $\mathbf X$ is transformed to $\mathbf X'$, which is written as
\begin {eqnarray}
\mathbf X' & = &  \begin{pmatrix} x'_1,& \cdots & x'_i, & \cdots & x'_N \end{pmatrix} \nonumber \\
& = & \mathbf M'_X\times 2^{ \varepsilon_\mathbf X} \nonumber
\end{eqnarray}
where
\begin {eqnarray}
\mathbf M'_{\mathbf X} & = &  \begin{pmatrix}m'_1,& \cdots & m'_i, & \cdots & m'_N \end{pmatrix} \nonumber \\
{ \varepsilon_\mathbf X} &=& \max_i{e_i|i\in[1,N]} \nonumber
\end{eqnarray}
$\varepsilon_X$ is the maximum exponent in the block $\mathbf X$ and $m_i$ is the aligned entry-wise mantissa that is derived with the following method,
\begin{equation}
m'_i = m_i >> ( \varepsilon_\mathbf X - e_i)
\label{mantissa}
\end{equation}
where $a>>b$ means right shifting $a$ with $b$ bits.

For CNN accelerator design, block-floating-point representation possesses two advantages. First, the concise expression contributes to saving the memory and the traffic bandwidth. If we have a block floating point format with $L_{e}$-bit exponent, $L_m$-bit mantissa,  and one sign bit, the average length of $n$ numbers is $1 + L_m + L_e / n$, while floating point representation costs $1 + L_m + L_e $ bits per number. The shorter averaged bit-width per number contributes to saving both of the memory and the off-chip bandwidth requirements.  In addition, with BFP, all multiply-accumulate operations in convolutional layer are carried out in fixed-point format. The fixed-point arithmetic unit in FPGA and ASIC design is much more efficient than the floating point one. For example, a 32-bit fixed-point adder in FPGA Virtex 7 690t consumes 1DSP with 300MHz clock speed. In contrast, a 16-bit 4-stage -pipeline floating-point adder is constituted of 2 DSPs and 117 LUT with 219MHz working frequency.

Acceleration in additions and multiplications is achieved at the cost of lower computation accuracy than floating-point counterpart because the small numbers in the block sacrificed  more valid bits during the block based aligning procedure as shown in equation (\ref{mantissa}). The errors during the BFP transform procedure are denominated as the quantization errors. There are two ways to handle the out-shifted bits, namely truncating and rounding off.
Our experiment proofed that rounding off outperforms truncating, because the truncation operation always generates the DC errors that can be accumulated in layer-wise and finally introduces a large bias.In contrast, the rounding operation introduces the zero-mean Gaussian white noises, and then no accumulated bias exists.

The energy of quantization errors of BFP is related to the distribution of numbers within the block, block size $n$ and mantissa bit length. To be specific, when $L_m$ is fixed, if $\mathbf X$ contains a few numbers with large magnitude while others are small, the overall precision of $\mathbf X'$ is low. When the distribution of numbers is given, the more numbers one block contains , the possibility of one block contains large peak and rather small mean value arises, resulting into a lower overall precision. Obviously, the precision of BFP is proportionate to $L_m$.

\subsection{Matrix Representation of Convolutional Neural Networks}
\begin{figure}
\begin{tikzpicture} [scale=0.6, every node/.style={scale=0.6}]\centering

\foreach \y in {-1,0}
\foreach \x in {6,7}
{
\draw (\x,\y) +(-.5,-.5) rectangle ++(.5,.5);

\draw (\x,\y) node{$w^0_{\pgfmathparse{(0 - \y)}\pgfmathprintnumber{\pgfmathresult},\pgfmathparse{ \x - 6} \pgfmathprintnumber{\pgfmathresult}} $};
}
\draw node at (6.5,-2) {kernel 0};

\foreach \y in {3,4}
\foreach \x in {6,7}
{
\draw (\x,\y) +(-.5,-.5) rectangle ++(.5,.5);

\draw (\x,\y) node{$w^1_{\pgfmathparse{(4 - \y)}\pgfmathprintnumber{\pgfmathresult},\pgfmathparse{ \x - 6} \pgfmathprintnumber{\pgfmathresult}}$};
}
\draw node at (6.5,2) {kernel 1};

\foreach \y in {0,1,2,3}
\foreach \x in {0,1,2,3}
{
\draw (\x,\y) +(-.5,-.5) rectangle ++(.5,.5);

\draw (\x,\y) node{$i_{\pgfmathparse{(3 - \y) } \pgfmathprintnumber{\pgfmathresult},\pgfmathparse{ \x - 0} \pgfmathprintnumber{\pgfmathresult}}$};
}
\draw node at (1.5,-1) {input feature map};
 \draw[dashed,gray,very thick]  (-0.6,3.7) -- (-0.6,1.4)  -- (1.6,1.4) -- (1.6,3.7) --(1.0,3.7) ;
 \draw node at (0.3,4) {receptive field};
\draw [scale = 1]node at (4.5, 1.5) {$\bigotimes$};
\foreach \y in {-1.5, -0.5, 0.5}
\foreach \x in {10,11,12}
{
\draw (\x,\y) +(-.5,-.5) rectangle ++(.5,.5);

\draw (\x,\y) node{$o^0_{\pgfmathparse{(0.5 - \y)}\pgfmathprintnumber{\pgfmathresult},\pgfmathparse{ \x - 10} \pgfmathprintnumber{\pgfmathresult}}$};
}
\draw node at (11,-2.5) {output feature map 0};
\draw [scale = 1]node at (8.5, -0.5) {$\Longrightarrow$};
\foreach \y in {2.5, 3.5, 4.5}
\foreach \x in {10,11,12}
{
\draw (\x,\y) +(-.5,-.5) rectangle ++(.5,.5);

\draw (\x,\y) node{$o^1_{\pgfmathparse{(5.5 - \y)}\pgfmathprintnumber{\pgfmathresult},\pgfmathparse{ \x - 10} \pgfmathprintnumber{\pgfmathresult}}$};
}
\draw node at (11,1.5) {output feature map 1};
\draw [scale = 1]node at (8.5, 3.5) {$\Longrightarrow$};

\end{tikzpicture}
\\
\begin{tikzpicture}[scale=0.6, every node/.style={scale=0.6},every delimiter/.style={scale=1.3}]\centering

\matrix (A) [matrix of math nodes , nodes = {},
	left delimiter  = (,%
             right delimiter = )] at (0,0)
{
\node {w^0_{0,0}}; & \node {w^0_{0,1}}; & \node {w^0_{1,0}}; & \node {w^0_{1,1}};  \\
\node {w^1_{0,0}}; & \node {w^1_{0,1}}; & \node {w^1_{1,0}}; & \node {w^1_{1,1}};  \\
};
 \draw[dashed, very thick, gray]  (-1.8, 1) -- (-1.8,0.1)  -- (1.8, 0.1) -- (1.8,1) --(-0.3,1) ;
 \draw node at (-1,1) {kernel 0};
\node [below] at (0,-1.2) {$\mathbf W$} ;
\matrix (B) [matrix of math nodes , nodes = {},
	left delimiter  = (,%
             right delimiter = )] at (7.5,0)
{
\node {i_{0,0}}; & \node {i_{0,1}}; & \node {i_{0,2}}; & \node {i_{1,1}}; &\node {i_{1,2}}; & \node {i_{2,0}}; & \node {i_{2,0}}; & \node {i_{2,1}}; & \node {i_{2,2}};  \\
\node {i_{0,1}}; & \node {i_{0,2}}; & \node {i_{0,3}}; & \node {i_{1,2}}; &\node {i_{1,3}}; & \node {i_{2,1}}; & \node {i_{2,1}}; & \node {i_{2,2}}; & \node {i_{2,3}};  \\
\node {i_{1,0}}; & \node {i_{1,1}}; & \node {i_{1,2}}; & \node {i_{2,1}}; &\node {i_{2,2}}; & \node {i_{3,0}}; & \node {i_{3,0}}; & \node {i_{3,1}}; & \node {i_{3,2}}; \\
\node {i_{1,1}}; & \node {i_{1,2}}; & \node {i_{1,3}}; & \node {i_{2,2}}; &\node {i_{2,3}}; & \node {i_{3,1}}; & \node {i_{3,1}}; & \node {i_{3,2}}; & \node {i_{3,3}}; \\
};
\node [below] at (7.5,-1.2) {$\mathbf I$} ;
 \draw[dashed,gray,very thick]  (4,1.2) -- (4,-1.2)  -- (4.9,-1.2) -- (4.9,1.2) --(4.7,1.2) ;
 \draw node at (4.8,1.5) {receptive field};
\matrix (C) [matrix of math nodes , nodes = {},
	left delimiter  =(,%
             right delimiter = )] at (4,-3)
{
\node {o^0_{0,0}}; & \node {o^0_{0,1}}; & \node {o^0_{0,2}}; & \node {o^0_{1,0}}; &\node {o^0_{1,1}}; & \node {o^0_{1,2}}; & \node {o^0_{2,0}}; & \node {o^0_{2,1}}; & \node {o^0_{2,2}}; \\
\node {o^1_{0,0}}; & \node {o^1_{0,1}}; & \node {o^1_{0,2}}; & \node {o^1_{1,0}}; &\node {o^1_{1,1}}; & \node {o^1_{1,2}}; & \node {o^1_{2,0}}; & \node {o^1_{2,1}}; & \node {o^1_{2,2}}; \\
};
\node [below] at (4,-3.7) {$\mathbf O$} ;
\draw node at (-2,-3) {$=$};
 \draw[dashed, very thick, gray]  (0.3, -2) -- (0.3,-3)  -- (7.7, -3) -- (7.7,-2) --(3.3,-2) ;
 \draw node at (1.8,-2) {output feature map 0};
\end{tikzpicture}
\caption{convolution operation transformed into matrix multiplication. ``$\mathbf W$'', ``$\mathbf I$'' and ``$\mathbf O$'' represent matrices transformed from kernels, input feature maps and output feature maps respectively. In this figure, the padding and stride are set to 0 and 1 with 1 channel.}
\label{fig:im2col}
\end{figure}
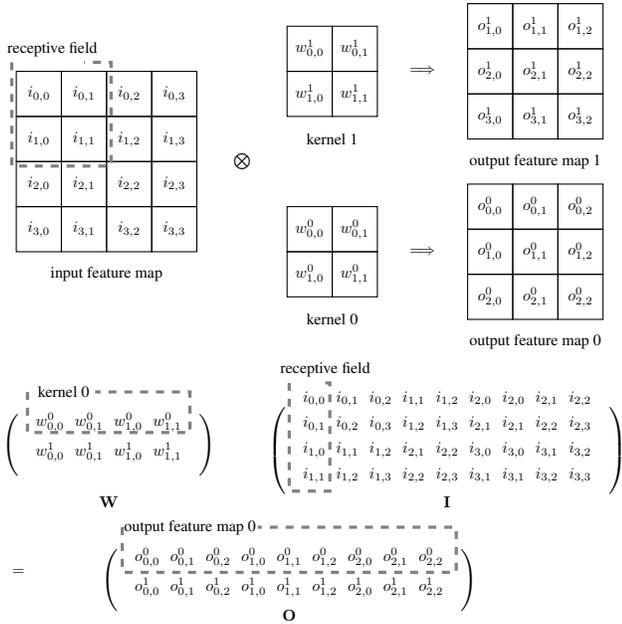

\noindent As transforming the convolution to matrix operation, kernels and input feature maps are expanded into two matrices namely $\mathbf W$ and $\mathbf I$. To be specific, kernels belonging to the same output feature map compose one row vector of $\mathbf W$, and receptive fields of input feature maps of one output pixel constitute one column vector in $\mathbf I$. This procedure is illustrated as figure \ref {fig:im2col}. The entry in $\mathbf O$ located at $m$th row, $n$th column corresponds to the output feature map of $m$th kernel on $n$th receptive field.It should be noted that, transforming CNN to matrix operation is burdensome. Therefore, the high performance CNN accelerators always apply the direct convolution data flow\cite{mei2017a200mhz}.In this paper, we merely adopt the matrix representation to explain  the BFP in CNN computation.

\subsection{Hardwired CNN Accelerator Oriented Matrix Partition for BFP Representation}
As aforementioned, block formatting $\mathbf W$ and $\mathbf I$ facilitates the advantages of BFP in hardwired CNN accelerator design. The precision of BFP is affected by the distribution of numbers within the block, the block size and the mantissa bit length. As the distribution of input feature maps and weights are predetermined, we can only optimize the other two factors, namely block size and mantissa bit length, to improve the overall performance. Under this guideline, the prominent issue is how to partition $\mathbf W$ and $\mathbf I$. The matrix multiplication is written as
\begin{equation}
\mathbf O_{M \times N} = \mathbf W_{M\times K}\mathbf I_{K\times N},
\label{all_all}
\end{equation}
where, $M$, $K$ and $N$ denote the number of output feature maps, the size of filters, and the size of one output feature map, respectively.
From the entry-wise perspective, matrix multiplication is represented as
 \begin{equation} \label {one_one}
o_{mn} = \vec w_{m}^T  \cdot \vec i_{n}
\end{equation}
and, if describled in row-wise or column-wise, it is recasted to
 \begin{equation} \label {one_all}
 \vec o_{m}^T = \vec w_{m}^T \cdot \mathbf I
\end{equation}
\begin{equation} \label {all_one}
\vec o_{n} = \mathbf W \cdot \vec i_{n}
\end {equation}

In fact, (\ref {all_all}), (\ref{one_one}), (\ref{one_all}) and (\ref{all_one}) illustrate four different ways to block format $\mathbf W$ and $\mathbf I$. Equation (\ref {all_all}) shows  $\mathbf W$ and $\mathbf I$ are block formatted as a whole respectively, thus the storage requirement reaches minimum at the price of the worst accuracy loss. (3) presents the vector-wise block formatting of $\vec w_m^T$ and $\vec i_n$, respectively. In this case, the minimum loss is achieved with increasing memory cost. Equation (\ref {one_all}) and (\ref{all_one}) represent two balanced approaches that obtained a good tradeoff between the quantization accuracy and the memory requirement, for they both block formats one matrix as a whole while the other one by row vector or column vector. The complexity and resource consuming comparisons of the above BFP transform methods are illustrated in Table \ref{tab:equation}.
\begin{table}[th]
\setlength\extrarowheight{4.5pt}
\resizebox{3.3in}{!}{%
\begin{tabular}{c l l l}

\hline
\hline
Method & $AL_{\mathbf W'}$ & $AL_{\mathbf I'}$ & NBE \\
\hline
Equation (\ref{all_all}) &  $1 + L_\mathbf W + L_e / (M \times K) $ & $1 + L_\mathbf I + L_e / (K \times N)$ & 2 \\
Equation (\ref{one_one}) &  $1 + L_\mathbf W + L_e /K$ & $1 + L_\mathbf I + L_e / K$ & M + N \\
Equation (\ref{one_all}) &  $1 + L_\mathbf W + L_e / K$ & $1 + L_\mathbf I + L_e / (K \times N)$ & 1 + M \\
Equation (\ref{all_one}) &  $1 + L_\mathbf W + L_e / (M \times K) $ & $1 + L_\mathbf I + L_e / K$ & 1 + N \\
\hline
\end{tabular}%
}
\caption{the cost of 4 different methods block formatting $\mathbf W_{M\times K}$ and $\mathbf I_{K\times N}$. ``$AL_{\mathbf W'}$'', ``$AL_{\mathbf I'}$'' are the average storing length of $\mathbf W'$ and $\mathbf I'$. ``NBE'' is the number of block exponents that need to store.}
\label{tab:equation}
\end{table}

Consider the layer "conv1\_1" of VGG-16, with the matrix representation of (2), we have $M = 64$, $K = 9$ and $N=50176$, where $N$ is much greater than $M$. According to table \ref{tab:equation}, equation (\ref{one_one}) and (\ref{all_one}) involve more than 50176 times of block formatting operation, besides, the cost of storing common exponents is hundreds of times ($50176 / 64$) larger than equation (\ref{all_all}) and (\ref{one_all}). The major difference of equation (\ref{all_all}) and (\ref{one_all}) is the block size of $\mathbf W$. We tested the influence of block size on accuracy, shown in table \ref{tab:sizeeffect}. Experiment revealed that the top-1 accuracy of equation (\ref{one_all}) is 1.6\% higher than equation(\ref{all_all}).Therefore, we choose equation(\ref{one_all}) to block format $\mathbf W$ and $\mathbf I$.
\begin{table} [th]
\setlength\extrarowheight{4.5pt}
\resizebox{3.3in}{!}{%
\begin{tabular}{l c c }
\hline
\hline
Method & Top-1 Accuracy & Top-5 Accuracy \\

Equation(\ref{all_all}) & 0.6672&  0.8768\\
Equation(\ref{one_all}) & 0.6832 & 0.884  \\
Floating point & 0.6808 & 0.8816\\
\hline
\end{tabular} %
}

\caption{The impact of block size on accuracy, tested in VGG-16 on dataset ILSVRC12\cite{ILSVRC15} with batch size set to 50. }
\label{tab:sizeeffect}
\end{table}
\subsection{Data Flow of Block Formatting in CNN}
\noindent For instance, it is given that $$\mathbf I = \begin{pmatrix} (1.01)_2 \times 2^0 &(1.01)_2 \times 2^0\\ (1.01)_2 \times 2^1&(1.01)_2 \times 2^2 \end{pmatrix}$$ $$\mathbf W  = \begin{pmatrix} (1.00)_2 \times 2^{-1}&(1.01)_2 \times 2^0 \end{pmatrix}$$
Let $L_\mathbf W = 3$, $L_\mathbf I = 3$ denominate the block mantissa bit length of $\mathbf W'$ and $\mathbf I'$, neglecting the sign bit. After scanning $\mathbf I$, we get the max exponent is $\varepsilon_\mathbf I = 2$, and then the entries in $\mathbf I$ are right shifted with round-off model to align.
Then, $$\mathbf I' = \begin{pmatrix} (0.01)_2 &(0.01)_2 \\ (0.11)_2&(1.01)_2  \end{pmatrix} \times 2^2$$It is traced by analogy that, $$\mathbf W'  = \begin{pmatrix} (0.10)_2&(1.01)_2  \end{pmatrix}\times 2^0$$ 
Therefore, the multiplication of $W$ and $I$, i.e., $O=WI$, can be approximated as
\begin{displaymath}
\mathbf O\approx \mathbf W' \mathbf I'=2^{\varepsilon_\mathbf O}\mathbf M'_\mathbf O
\end{displaymath}
where, $\mathbf M'_\mathbf O=\mathbf M'_\mathbf W\mathbf M'_\mathbf I$ and $\varepsilon_\mathbf O=\varepsilon_\mathbf W+\varepsilon_\mathbf I$.

\begin{figure*}
\centering
\begin{tikzpicture} [scale=0.67, every node/.style={scale=0.7}] \large
	\tikzstyle{adder}=[circle,draw, thick]
	\tikzstyle {multiplier} = [circle,draw,thick]
	\tikzstyle {source} = [circle,draw,thick]

	\draw [color=black,->,thick](7,10) node[above]{$\mathbf W_{R_i}$}-> (7,9);
	\draw [color=black, thick] (6,8) rectangle (8, 9)  (7,8.5)node [color=black!100,thick]{FP2BFP};
	
	\draw [color=black,->,thick](4,10) node[above]{$\mathbf I$}-> (4,9);
	\draw [color=black, thick] (3,8) rectangle (5, 9)  (4,8.5)node [color=black!100,thick]{FP2BFP};
	
	\draw[color=black,thick, ->](4,8) -- (4,6.3)->(4.3,6.3);
	\draw[color=black,thick, ->](7,8) -- (7,6.3) ->(6.7, 6.3);
	\draw [color=black, thick] (4.3,5.3) rectangle (6.7, 7.3)  (5.5,6.6)node [color=black!100,thick][above]{Fixed point} (5.5,6.6)node [color=black!100,thick][below]{ matrix } (5.5,6.1)node [color=black!100,thick][below] {multiplication};
	
	\draw[color=black,thick, ->](5.5, 5.3) -- (5.5, 4.3);
	\draw[color=black, thick] (4.5,2.8) rectangle (6.5, 4.3)  (5.5,3.55)node [color=black!100,thick]{BFP2FP};
	
	\draw[color=black,thick, ->](5.5, 2.8) -- (5.5, 1.8) node [below] [color=black!100,thick] {$\mathbf O$};
	
	\draw [color=black,->,thick](0,10) node[above]{$\mathbf P$}-> (0,9.5);
	\draw [color=black, thick] (-2,8.5) rectangle (2, 9.5)  (0,9)node [color=black!100,thick]{Derive $\varepsilon_{\mathbf P}$};
	
	\draw [color=black,->,thick](0,8.5) -- (0,8);
	\draw [color=black, thick] (-2,7) rectangle (2,8)  (0,7.5)node [color=black!100,thick]{Block based aligning};
	
	\draw [color=black,->,thick](0,7) -- (0,6.5);
	\draw [color=black, thick] (-2,5.5) rectangle (2,6.5)  (0,6)node [color=black!100,thick]{Round off};
		
	\draw[color=black,thick, ->](0, 5.5) -- (0,5) node [below] [color=black!100,thick] {$\mathbf P'$};
	
	\draw (-1, 10.5) node {FP2BFP};
	\draw[dashed] (0,10.5) -- (2.5, 10.5) -- (2.5, 4.5) -- (-2.2,4.5) --(-2.2, 10.5) --(-2,10.5);

	\draw [color=black,->,thick](0,3.5) node[above]{$\mathbf P'$}-> (0,3);
	\draw [color=black, thick] (-2,2) rectangle (2, 3)  (0,2.5)node [color=black!100,thick]{$\mathbf P = \mathbf M'_\mathbf P \cdot 2 ^ {\varepsilon_{\mathbf P}}$};
	\draw [color=black,->,thick](0,2) -> (0,1.5)node[below]{$\mathbf P$};
	\draw (-1, 4.2) node {BFP2FP};
	\draw[dashed] (0,4.2) -- (2.5, 4.2) -- (2.5, 1) -- (-2.2,1) --(-2.2, 4.2) --(-2,4.2);
	
	\node [adder] (adder) at (12.75, 3.5) {$+$};
	\node (output) at  (12.75, 2) {$o_{ij}$};
		\node [multiplier] (m1) at (9.75,7) {$\times$};
		\node [multiplier] (m2) at (12.75,7) {$\times$};
		\node [multiplier] (m3) at (15.75,7) {$\times$};
	
		  \draw [color=black,->,thick](9.75,6.6)  -- node[left]{$   [L_{\mathbf W} + L_{\mathbf I} - 1:0]$}  ++   (adder);
		  \draw [color=black,->,thick](12.75,6.6)  --   (adder); 
		  \draw [color=black,->,thick](15.75,6.6)  --  (adder); 
	\node  (a1)  at (9, 9) {$i_{{i1}}'$};
	\node (b1) at (10.5, 9) {$w_{{1j}}'$};
	\draw [color=black,->,thick](a1)  --   (m1);
	\draw [color=black,->,thick](b1)  --  (m1);
	
	\node  (ak)  at (12, 9) {$i_{{ik}}'$};
	\node (bk) at (13.5, 9) {$w_{{kj}}'$};
	\draw [color=black,->,thick](ak)  -- node[left]{$   [L_{\mathbf I}-1:0]$}  ++ (m2); 
	\draw [color=black,->,thick](bk)  -- node[right]{$   [L_{\mathbf W}-1:0]$}  ++ (m2);	
	
	\node  (aK)  at (15, 9) {$i_{{iK}}'$};
	\node (bK) at (16.5, 9) {$w_{{Kj}}'$};
	\draw [color=black,->,thick](aK) --  (m3) ; 
	\draw [color=black,->,thick](bK) --  (m3); 
	
      \draw[color=black,->,thick](adder) -- node[right]{$   [L_{\mathbf W} + L_{\mathbf I} -1+ S :0]$}  ++ (output) ;
      \draw (12, 10) node {Fixed point matrix multiplication};
      \draw[dashed]  (9, 10) -- (8.1,10)  -- (8.1, 1) -- (17.2,1) --(17.2,10) --(15,10);

\end{tikzpicture}
\caption{theoretical data flow of block floating point. Weight matrix and input matrix are block formatted individually and then matrix multiplication is done via fixed-point accumulators and multipliers. In this figure, $L_{\mathbf I}$ and $L_{\mathbf W}$ both includes the sign bit.}
\label{fig:dataflow}
\end{figure*}
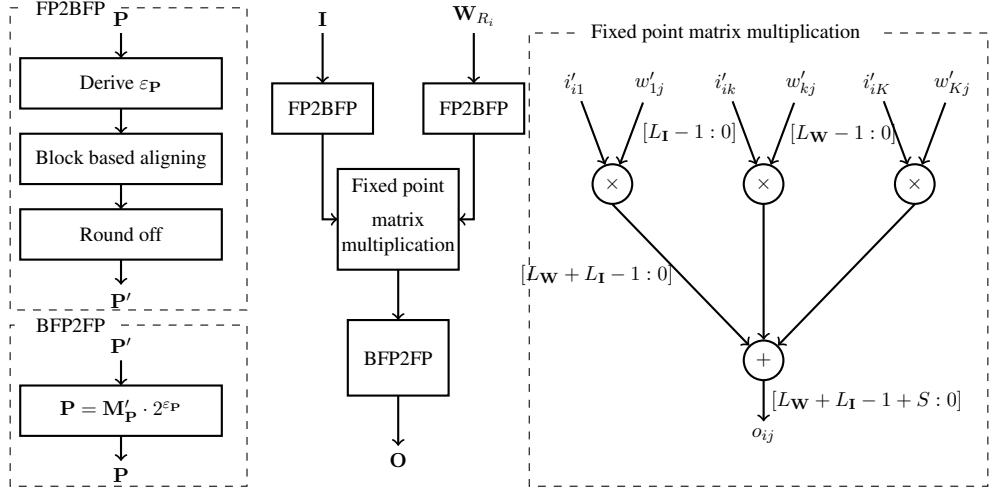

To avoid involving rounding errors during $\mathbf M'_{\mathbf O} = \mathbf  M'_{\mathbf W} \mathbf M'_{\mathbf I}$, the bit width of multiplier must be no less than $L_\mathbf W + L_\mathbf I + 2$, including the sign bit, and the bit width of accumulator must be no less than $L_\mathbf W + L_\mathbf I + 2 + S$, where $S = \lfloor log_2(K)\rfloor$ to prevent overflow as $K$ times binary addition generates $\lfloor log_2(K)\rfloor$ times carry at most. Details are shown in figure \ref{fig:dataflow}.

\section{Error Analysis of Block Floating Point Convolution Operations}
\noindent We propose a three-stage error analysis model. The first stage is the quantization error, the second stage describes the procedure of error accumulation in matrix multiplication, and the third one describes how the errors are transported  between convolution layers. 
\subsection{Quantization Error Analysis Model}
According to \cite{kalliojarvi1996roundoff}, for block $\mathbf X$, the quantization error has zero mean , and variance $\sigma^2$
\begin{equation}\label{noise_variance}
\sigma^2 = \frac{2^{-2L_{m}}}{12}\cdot\sum_{i=1}^{N_\gamma}p_{\gamma_i}2^{2\gamma_i}
\end{equation}
where $L_{m}$ is the bit length of block mantissa and $p_{\gamma_i}(i = 1, \cdots, N_{\gamma})$ is the probability mass function (PMF) of the block-exponents. $N_{\gamma} = 2^{L_E}$ is the number of available block-exponent levels, where $L_{E}$ is the bit length of block exponent.

As the value of input feature maps and weight filters are known, $p_{\gamma_i}$ is described as below,
\begin{equation}
p_{\gamma_i} =
\begin{cases}
1 \quad {i = \varepsilon_{\mathbf X}}\\
0 \quad {i \neq  \varepsilon_{\mathbf X}}
\end{cases}
\label{7}
\end{equation}
Substituting (\ref{7}) to (\ref {noise_variance}), we derive that
\begin{equation} \label{noise_variance_ex}
\sigma_{\alpha}^2 = \frac{2^ {-2L_{m}}}{12}\cdot 2^{2\cdot{ \varepsilon_{\mathbf X}}}
\end{equation}
Based on equation (\ref{one_all}),  the input matrix is treated a $K\times N$ block as a whole and the weight matrix is partitioned into $M$ numbers of $1\times K$ row vectors. Thus the signal-to-noise ratio (SNR) of block floating point represented input matrix is
\begin{eqnarray} \label{SNR_input}
SNR_{i} =10\cdot\log_{10}\frac{E(Y^2)}{\sigma_{i}^2}
\end{eqnarray}
where $E(Y^2)$ is the mean square of input matrix, $\sigma_{i}^2$ is the energy of quantization error of $\mathbf I'$. To be specific,
\begin{eqnarray}
\sigma_{i}^2 = \frac{2^ {-2L_{\mathbf I}}}{12}\cdot 2^{2\cdot \varepsilon_{\mathbf I}}
\end{eqnarray}

Similarly, SNR of the $m$th BFP represented row vector in the weight matrix is
\begin{equation} \label{mweight}
SNR_{wm} = 10\cdot\log_{10}\frac{E(X_{m}^2)}{\sigma_{wm}^2}
\end{equation}
where $E(X_{m}^2)$ is the mean square of the $m$th row vector of weight matrix and $\sigma_{wm}^2$  the corresponding energy of quantization errors that is formulated as,
\begin{eqnarray}
\sigma_{wm}^2 = \frac{2^ {-2L_{\mathbf W}}}{12}\cdot 2^{2\cdot { \varepsilon_{ \vec w_{m}^T}}}
\end{eqnarray}
The averaged SNR of whole weight matrix is
\begin{equation} \label{SNR_weight}
SNR_{w} = 10\cdot\log_{10}{\frac{\sum_{m=1}^{M}{E(X^2_m)}}{\sum_{m=1}^{M} \sigma_{wm}^2}}
\end{equation}
\subsection {Single Layer Error Analysis Model}
\noindent Matrix multiplication is composed of vector inner products. Therefore, investigating the vector inner product assists us in understanding how error is accumulated in BFP represented matrix multiplication.
Giving two vectors with length $K$ as $\vec P$ and $\vec Q$, which are block formatted into $\vec P_b$ and $\vec Q_b$. We further define $\vec P_e = \vec P_b - \vec P$ and $\vec Q_e = \vec Q_b - \vec Q$ as the quantization errors. Then the mean square of block floating point represented inner product $\sigma_r^2$ is
\begin {eqnarray}
\sigma_r^2&=&E((\vec P_b\cdot \vec Q_b)^2) \nonumber \\
&=&E((\vec P\cdot \vec Q)^2) + E((\vec P_e \cdot \vec Q)^2) + \nonumber \\
&& E((\vec P \cdot \vec Q_e)^2) + E((\vec P_e \cdot \vec Q_e)^2)
\end{eqnarray}
Assuming that $\vec P_e$ and $\vec Q_e$ are statistically independent, and ignoring the higher order item $E((\vec P_e \cdot \vec Q_e)^2)$ , we have
\begin {eqnarray} \label {vector_model}
\sigma_r^2&=&E((\vec P\cdot \vec Q)^2) + E((\vec P_e \cdot \vec Q)^2) +E((\vec P \cdot \vec Q_e)^2) \nonumber \\
&=&\frac{1}{K}(1+\frac{\| \vec P_e\| ^ 2}{\| \vec P \| ^2} + \frac{\| \vec Q_e \| ^2}{\| \vec Q\| ^2})
\cdot \| \vec P\| ^2 \cdot \| \vec Q\| ^2
\end{eqnarray}
where $$\| \vec P \| ^2 = \sum_{k = 1}^{K}P_i^2, \| \vec Q\| ^2 = \sum_{k = 1}^{K}Q_i^2$$
$\frac{\| \vec P_e\| ^ 2}{\| \vec P \| ^2}$ and $\frac{\| \vec Q_e \| ^2}{\| \vec Q\| ^2}$, denoted as $\eta_P$ and $\eta_Q$, are noise-to-signal-ratio (NSR) of $\vec P_b$ and $\vec Q_b$, which can be derived from SNR, e.g.
\begin{equation} \label{NSR_SNR}
 \eta_P = 10 ^{-\frac{SNR_P}{10}} \nonumber
\end{equation}
where ${SNR_P}$ has been discussed in equation (\ref{SNR_input}). Then the NSR of inner product is
\begin{eqnarray}
\eta_r & = & \frac{\sigma_r^2 - E((\vec P\cdot \vec Q)^2) }{E((\vec P\cdot \vec Q)^2)} \nonumber \\
& = & \eta_P + \eta_Q
\end{eqnarray}

Since $o_{mn} = \vec w^T_{m} \cdot \vec i_{n}$, we can use equation (\ref{vector_model}) to calculate its NSR. Further, when calculating the average NSR of $\mathbf O$, we assume that $\vec w^T_{m}$ are independent and identically distributed, similarly to $\vec i_{n}$, then NSR of $\vec w^T_{m}$ and  $\vec i_{n}$ can be replaced with the NSR of $\mathbf W'$ and $\mathbf I'$. Thus the average NSR of $\mathbf O$, denoted as $\eta_{\mathbf O}$, is
\begin{equation}
\eta_{\mathbf O} = \eta_{\mathbf I'} + \eta_{\mathbf W'}
\end{equation}
where $ \eta_{\mathbf I'}$ and $ \eta_{\mathbf W'}$ are NSR of input matrix and weight matrix. Substituting equation (\ref{NSR_SNR}), SNR of output matrix is
\begin {eqnarray} \label{single_layer_model}
SNR_{\mathbf O}& = &-10\cdot \log_{10}{\eta_{o}} \nonumber \\
&=&SNR_{\mathbf I'} + SNR_{\mathbf W'} - 10 \cdot \log_{10} \nonumber \\
&&{(10^{\frac{SNR_{\mathbf I'}}{10}} + 10^{\frac{SNR_{\mathbf W'}}{10}}) }
\end {eqnarray}
where $SNR_{\mathbf I'}$ and $SNR_{\mathbf W'}$ have been discussed in equation (\ref{SNR_input}) and (\ref{SNR_weight}), thus we get the single layer error analysis model as equation (\ref{single_layer_model}).

\subsection {Multi-Layers Error Analysis Model}
\noindent In VGG-16,  every convolution layer is followed by a ReLU layer, and the output of ReLU is the input of next convolution layer. To simplify our model, we assume that the errors are uniformly distributed in negative and positive output feature maps, and then we ignore the impact of ReLU layer on SNR. The difference between multi-layers model and single layer model is that the original input feature maps of multi-layers model carries error while the single layer does not. Fortunately, the quantization errors are uniformly distributed in the input signals and the input inherited errors. Hence, we can utilize single layer model to  calculate the new generated error, and then we use the SNR of last layer to distinguish the carried error and signal.

 $\eta_{1}$ and $\eta_{2}$ stand for the last layer output NSR and the NSR of block formatted input feature maps. $E(Y^2)$, $\sigma_1^2$ and $\sigma_2^2$ are the energy of signal, the energy of error inherited from the last layer and the energy of quantization error. Based on  equation (\ref{SNR_input}) and (\ref{NSR_SNR}) ,
\begin{equation}
\eta_2 = \frac{\sigma_2^2}{E(Y^2) + \sigma_1^2}
\end{equation}
where $\sigma_1^2 = \eta_{1}\cdot E(Y^2) $ and $\sigma_2^2$ are derived from equation (\ref{noise_variance_ex}).  And then, the overall NSR $\eta$ of this input feature map is
\begin{eqnarray}
\eta &=& \frac{\eta_2(E(Y^2) + \eta_1E(Y^2))}{E(Y^2)} \nonumber \\
& = &\eta_2 + \eta_1\eta_2
\end{eqnarray}

\begin{table*}[htbp]\large
\setlength\extrarowheight{7pt}
\resizebox{7in}{!}{%
\begin{tabular}{c c c c c c c c c c c c c c c c c c c c}
\hline
\hline
\multicolumn{5}{c}{VGG-16 top-1} & \multicolumn{5}{c}{GoogLeNet loss1 top-1} & \multicolumn{5}{c}{GoogLeNet loss2 top-1} & \multicolumn{5}{c}{GoogLeNet loss3 top-1}\\
\multirow{2}{*}{$L_{\mathbf W}$} & \multicolumn{4}{c}{$L_{\mathbf I}$} &  &  \multicolumn{4}{c}{$L_{\mathbf I}$} & & \multicolumn{4}{c}{$L_{\mathbf I}$} &  &  \multicolumn{4}{c}{$L_{\mathbf I}$}  \\
\cline{2-5}
\cline{7-10}
\cline{12-15}
\cline{17-20}
& 6 & 7 & 8 & 9 & & 6 & 7 & 8 & 9 & & 6 & 7 & 8 & 9 & & 6 & 7 & 8 & 9 \\
6 & 0.3096  &  0.1576  &  0.1246  &  0.12 & & 0.022  &  0.0126  &  0.0122  &  0.0096 && 0.0198 & 0.0138 & 0.0118 & 0.01 & & 0.0272 & 0.0094 & 0.0088 & 0.0072 \\
7 & 0.185  &  0.0268  &  0.003  &  0.0022  & & 0.0102  &  0.0004  &  0.0014  &  0.0012 && 0.012 & 0.004 & 0.0014 & 0.0008  & & 0.0172	 & 0.0028 & 0.0014 & -0.0004 \\
8 & 0.1772  &  0.0168  &\cellcolor{Gray}  0.0002  &\cellcolor{Gray}  -0.0008 & & 0.0036  &  	-0.0012  & \cellcolor{Gray} -0.0008  & \cellcolor{Gray} -0.0004 && 0.0156 & 0.004 &\cellcolor{Gray} 0.0008 &\cellcolor{Gray} 0.0008  & & 0.017 & 	0.0064 &\cellcolor{Gray} 0.0014 &\cellcolor{Gray} 0.003 \\
9 & 0.1764  &  0.0166  &\cellcolor{Gray}  -0.0002  &\cellcolor{Gray}  -0.0018 & & 0.0078  &  	-0.002  &\cellcolor{Gray}  -0.0004  &\cellcolor{Gray}  -0.0012 &&0.014 & 0.0002 &\cellcolor{Gray} 0.0018 &\cellcolor{Gray} 0.0008  & & 0.014 & 0.0032 &\cellcolor{Gray} 0.0004 &\cellcolor{Gray} 0.0012 \\
\hline
\hline
\multicolumn{5}{c}{ResNet-18 top-1} & \multicolumn{5}{c}{ResNet-50 top-1} & \multicolumn{5}{c}{mnist} & \multicolumn{5}{c}{cifar10}\\
\multirow{2}{*}{$L_{\mathbf W}$} & \multicolumn{4}{c}{$L_{\mathbf I}$} &  &  \multicolumn{4}{c}{$L_{\mathbf I}$} &\multirow{2}{*}{$L_{\mathbf W}$}& \multicolumn{4}{c}{$L_{\mathbf I}$} & \multirow{2}{*}{$L_{\mathbf W}$}  &  \multicolumn{4}{c}{$L_{\mathbf I}$}   \\
\cline{2-5}
\cline{7-10}
\cline{12-15}
\cline{17-20}
& 6 & 7 & 8 & 9 & & 6 & 7 & 8 & 9 & &3 & 4 & 5 & 6 & & 5 & 6 & 7 & 8 \\
6 & 0.184 & 0.0584 & 0.0518 & 0.0506 &  & 0.1038 & 0.0348	 & 0.0224 & 0.0186 & 3 & 0.0123 & 0.0068 & 0.0053 & 0.0045 & 5 & 0.0219 & 	0.0103 & 	0.0105 & 0.0087\\
7 & 0.125 & 0.019 & 0.008 & 0.0052 &  & 0.0724 & 0.0128 & 0.0064	 & 0.0024 & 4 & 0.0051 &\cellcolor{Gray} 0.0010 & 0.0005  & -0.0002 & 6 & 0.0145 & 0.0034 & 0.0014 & 0.0015\\
8 & 0.1228 & 0.012 &\cellcolor{Gray} 0.0026 &\cellcolor{Gray} 0 &  & 0.0664 & 0.0074 & \cellcolor{Gray}0.0008 &\cellcolor{Gray} -0.0022 & 5 & 0.0054 & 0.0006 &\cellcolor{Gray} 0.0001 & -0.0002  & 7 & 0.0169 & 0.0042 &\cellcolor{Gray} 0.0028 &\cellcolor{Gray} 0.0014\\
9 & 0.1134 & 0.01 &\cellcolor{Gray} -0.0006 &\cellcolor{Gray} 0 &  & 0.058 & 0.0084	 & \cellcolor{Gray}0.0028 &\cellcolor{Gray} 0.0004 & 6 & 0.0051 & 0.0010  & 0.0004 & -0.0005 & 8 & 0.0166 & 0.0014 &\cellcolor{Gray} 0.002 &\cellcolor{Gray} -0.0009\\
\hline
\end{tabular}%
}
\caption{Drop of accuracy in VGG-16, GoogLeNet, ResNet-18, ResNet-50 ,cifar10 and mnist. $L_{\mathbf W}$ and $L_{\mathbf I}$ represent the block mantissa bit length ( including the sign bit ) of $\mathbf W'$ and $\mathbf I'$ respectively.}
\label{accuracy}
\end{table*}

\subsection {Deviation of Error Analysis Model}
\subsubsection {Correlation between Filters and Input Feature Maps}
\noindent We assumed that weights and input feature maps are statistically independent to simplify our single layer error analysis model. However, when weights and input feature maps are rather strong correlated, which results into SNR arising as noise is independent to weights while signal is not. In this case, our model deviates from it. Another indication of strong correlation is that strong correlated layers generate more large values compared with others as filters extract aimed features from receptive fields. The higher the degree of coincidence is tends to generates more large values.
\subsubsection {ReLU Layer}
\noindent ReLU\cite{glorot2011deep} is a nonlinearity layer, which drops values smaller than zero and keeps positive values as they are. In VGG-16, each convolution layer is followed by a ReLU layer, of which the outputs are dispatched to the following  convolution layer or max pooling layer. In our multi-layers model, we used  SNR of last convolution layer's output as SNR of next convolution layer's input matrix, thus the influence of ReLU layer is ignored. Further,  our model works for any activation function whose derivate is descending, because their output NSR is always no greater than input NSR \cite{7547305} (we recommend readers to read lemma 1 this literature it for more detailed proof).
\subsubsection {Pooling Layer}
\noindent VGG-16 uses max pooling layer every several convolution layers to lessen the number of parameters and to control overfitting. A max pooling layer extracts the biggest number of $2\times 2$ receptive filter with stride 2. It seems reasonable to assume that pooling layer always promote the overall SNR, if we assume bigger magnitude is sum of the products of  bigger multiplier, and because bigger magnitudes have higher SNR when represented in block floating point, the SNR of the biggest number with the   $2\times 2$ filter is higher than the average SNR of the filter. However, this does not necessarily be true as it is possible that big positive and negative magnitudes offset each other, resulting a rather small value, while smaller magnitudes accumulated to a big one that is selected as the output.  Because of the uncertainty pooling layer's impact on SNR, we take the output SNR of pooling layer as the input SNR of next layer.

\section {Experiments}
\subsection {Accuracy Verification of BFP CNN }
\noindent The magnitude of the decrease in accuracy is one of the most important criteria for measuring the performance of CNN accelerators. We verified BFP arithmetic on several typical deep neural networks, including VGG-16, GoogLeNet, ResNet-18 and ResNet-50, besides, smaller convolution neural networks like mnist and cifar10 are also tested.
\subsubsection{Experiment Setup}
\noindent Caffe\cite{jia2014caffe} is a popular deep learning framework, which turns convolution operations to matrix multiplications. It is convenient to apply BFP in CNN based on Caffe as we only need to rewrite the convolution function in caffe under the instruction of figure \ref{fig:dataflow}. To be specific, input feature maps and weights are block formatted accordingly, and then matrix multiply, finally the output feature map is transformed to floating point representation as $\mathbf O'$ holds different block exponent for different row vector, because weights are block formatted row by row.
It should be pointed out that ReLU and pooling layers remained unchanged, but this has no impact on our test as these two layers do not involve numeric computation.

\subsubsection{Results}
\noindent Results are shown in Table \ref{accuracy}. $L_{\mathbf W}$ and $L_{\mathbf I}$ denote the bit length of weight and input mantissa after block formatted, including sign bit. For deep neural networks, when set $L_{\mathbf W}$ and $L_{\mathbf I}$ no less than 8, the drop of accuracy is less than 0.3\%. In addition,  4-bit mantissa and 7-bit mantissa are sufficient for mnist and cifar10 respectively. In the experiments, we used the original models without any retraining, and then block formatted them with different mantissa length respectively. Thus the accuracy differences are introduced by the quantization errors merely.

Another noteworthy is that the decrease of accuracy is more sensitive to $L_{\mathbf I}$ than $L_{\mathbf W}$. This is attributed to two factors, namely the block size of $\mathbf I'$ is much larger than the size of $\mathbf W'$, and the dynamic range of input feature map is much larger than that in weights.


To draw a conclusion, when designing FPGA based CNN accelerators, BFP is a superduper numeric format as BFP eliminates the complex floating-point computations in convolution operation, while maintaining the high classification accuracy. Further, because BFP oriented accelerator does not acquire retraining, the cost of implementing BFP is low. Our experiments revealed that BFP can be used in a variety of convolution neural networks without specific reconfiguration.

\begin{table}[!]
\centering
\small
\resizebox{3.1in}{!}{%
\begin{tabular} {c c c c c  }
\hline
\hline
\multicolumn {2}{c}{Layer} & ex SNR & single SNR & multi SNR \\
\hline
\multirow{4}{*}{conv1\_1}
& input & 40.1236 & 41.8047 & --  \\
& weight & 43.9925 & 44.3538 & --  \\
&  \cellcolor{Gray} output & \cellcolor{Gray} 37.5638 &  \cellcolor{Gray} 39.8845 &  \cellcolor{Gray}  --  \\
& ReLU & 37.5641 & -- & -- \\
\hline
\multirow{4}{*}{conv1\_2}
& input & 27.2022 & 26.9376 & 26.7227  \\
& weight & 36.5345 & 37.3569 & 37.3569  \\
& \cellcolor{Gray} output & \cellcolor{Gray} 35.1682 & \cellcolor{Gray} 26.5601 & \cellcolor{Gray}  26.3628  \\
& ReLU & 35.1707 & -- & -- \\
\hline
pool1 & max & 36.3581 & -- & -- \\
\hline
\multirow{4}{*}{conv2\_1}
& input & 27.5767 & 29.3567 & 28.5668  \\
& weight & 34.1054 & 35.347 & 35.347  \\
& \cellcolor{Gray} output & \cellcolor{Gray} 30.0439 & \cellcolor{Gray} 28.3815 & \cellcolor{Gray} 27.7393 \\
& ReLU & 30.0446 & -- & -- \\
\hline
\multirow{4}{*}{conv2\_2}
& input & 23.7616 & 25.7545 & 23.6242  \\
& weight & 33.7565 & 34.9562 & 34.9562  \\
& \cellcolor{Gray} output & \cellcolor{Gray} 25.3109 & \cellcolor{Gray} 25.2616 & \cellcolor{Gray} 23.3158  \\
& ReLU & 25.311 & -- & -- \\
\hline
pool2 & max & 26.2151  & -- & --\\
\hline
\multirow{4}{*}{conv3\_1}
& input & 23.9214 & 27.9558 & 23.9885  \\
& weight & 31.3016 & 32.899 & 32.899  \\
& \cellcolor{Gray} output & \cellcolor{Gray}  25.2734 & \cellcolor{Gray} 26.7488 & \cellcolor{Gray} 23.4634 \\
& ReLU & 25.2733 & -- & -- \\
\hline
\multirow{4}{*}{conv3\_2}
& input & 21.4743 & 24.109 & 20.7639  \\
& weight &30.7485 & 32.1746 & 32.1746  \\
& \cellcolor{Gray} output & \cellcolor{Gray} 23.1478 & \cellcolor{Gray} 23.479 &  \cellcolor{Gray} 20.4609\\
& ReLU & 23.1478 & -- & -- \\
\hline
\multirow{4}{*}{conv3\_3}
& input & 20.1885 & 24.2099 & 18.9325  \\
& weight & 29.8594 & 31.3544 & 31.3544  \\
& \cellcolor{Gray} output & \cellcolor{Gray} 21.0608 & \cellcolor{Gray} 23.4435 & \cellcolor{Gray} 18.6907 \\
& ReLU & 21.0608 & -- & -- \\
\hline
pool3 & max & 21.7996 & -- & -- \\
\hline
\multirow{4}{*}{conv4\_1}
& input & 20.7986 & 25.7334 & 20.3252  \\
& weight & 31.0773 & 32.5038 & 32.5038  \\
& \cellcolor{Gray} output & \cellcolor{Gray} 22.9078 & \cellcolor{Gray} 24.9042 & \cellcolor{Gray} 20.0699 \\
& ReLU & 22.9077 & -- & -- \\
\hline
\multirow{3}{*}{conv4\_2}
& input & 19.3041 & 23.882	& 18.5602  \\
& weight & 31.0578 & 32.3566 & 32.3566  \\
& \cellcolor{Gray} output & \cellcolor{Gray} 21.9051 & \cellcolor{Gray} 23.305	& \cellcolor{Gray} 18.3827  \\
& ReLU & 21.9049 & -- & -- \\
\hline
\multirow{4}{*}{conv4\_3}
& input & 18.2669 & 24.0675 & 17.3443  \\
& weight & 30.2625 & 31.6326 & 31.6326  \\
& \cellcolor{Gray} output & \cellcolor{Gray} 22.4316 & \cellcolor{Gray} 23.3665 &  \cellcolor{Gray} 17.1855  \\
& ReLU & 22.4312 & -- & -- \\
\hline
pool4 & max & 18.8514 & -- & -- \\
\hline
\multirow{4}{*}{conv5\_1}
& input & 18.5113 & 24.4103 & 17.786  \\
& weight & 31.0754 & 32.242 & 32.242  \\
& \cellcolor{Gray} output & \cellcolor{Gray} 22.149 & \cellcolor{Gray} 23.7479 & \cellcolor{Gray} 17.6331 \\
& ReLU & 22.1483 & -- & -- \\
\hline
\multirow{4}{*}{conv5\_2}
& input & 18.4841 & 23.5987 & 16.6529  \\
& weight & 33.0316 & 33.9193 & 33.9193\\
& \cellcolor{Gray} output & \cellcolor{Gray} 22.2687 & \cellcolor{Gray} 23.2129 & \cellcolor{Gray} 16.5772  \\
& ReLU & 22.2369 & -- & -- \\
\hline
\multirow{4}{*}{conv5\_3}
& input & 18.1074 & 24.1601 & 15.8788 \\
& weight & 32.4689 & 33.654 & 33.654  \\
& \cellcolor{Gray} output & \cellcolor{Gray} 23.6306 & \cellcolor{Gray} 23.6976 & \cellcolor{Gray} 15.7846\\
& ReLU & 23.6191 & -- & -- \\
\hline
pool5 & max & 17.7955 & -- & -- \\
\hline
\end{tabular}%
}

\caption{Experimental and theoretical SNR. In this table, ``ex SNR'', ``single SNR'' and ``multi SNR'' respectively represent experimental SNR, single layer model calculated SNR and multi-layer model calculated SNR.}
\label{table:error}
\end{table}
\subsection{Error Analysis Model Verification}
\subsubsection {Experiments Setup}
\noindent To verify error analysis model, we defined floating point represented  numbers as signals, and the differences between floating point represented numbers and BFP represented numbers as errors. And then, we ran VGG-16 on ILSVRC2012 for 20 iterations with batch size set to 50 to gather data, such as the output of every layer and the input feature maps and weights of convolution layer. These data are stored in separated files in binary format, with which we calculate the signal energy and error energy to derive the experimental SNR.

\subsubsection {Results} As shown in Table \ref{table:error}, the theoretical analysis agrees well with  the experimental data, where the biggest difference between them is less than 8.9dB, which is close enough to guide hardware design. What worth to mention is that the previous assumptions about ReLU layer is proved to be reasonable. To be specific, the SNR of ReLU output is consistent with its input SNR, which proved that the output of convolution layer is evenly distributed in the positive and negative regions. And, the impact on SNR of pooling layer performs exactly as what we assumed.

\begin{figure}
\centering
\includegraphics[scale = 0.5]{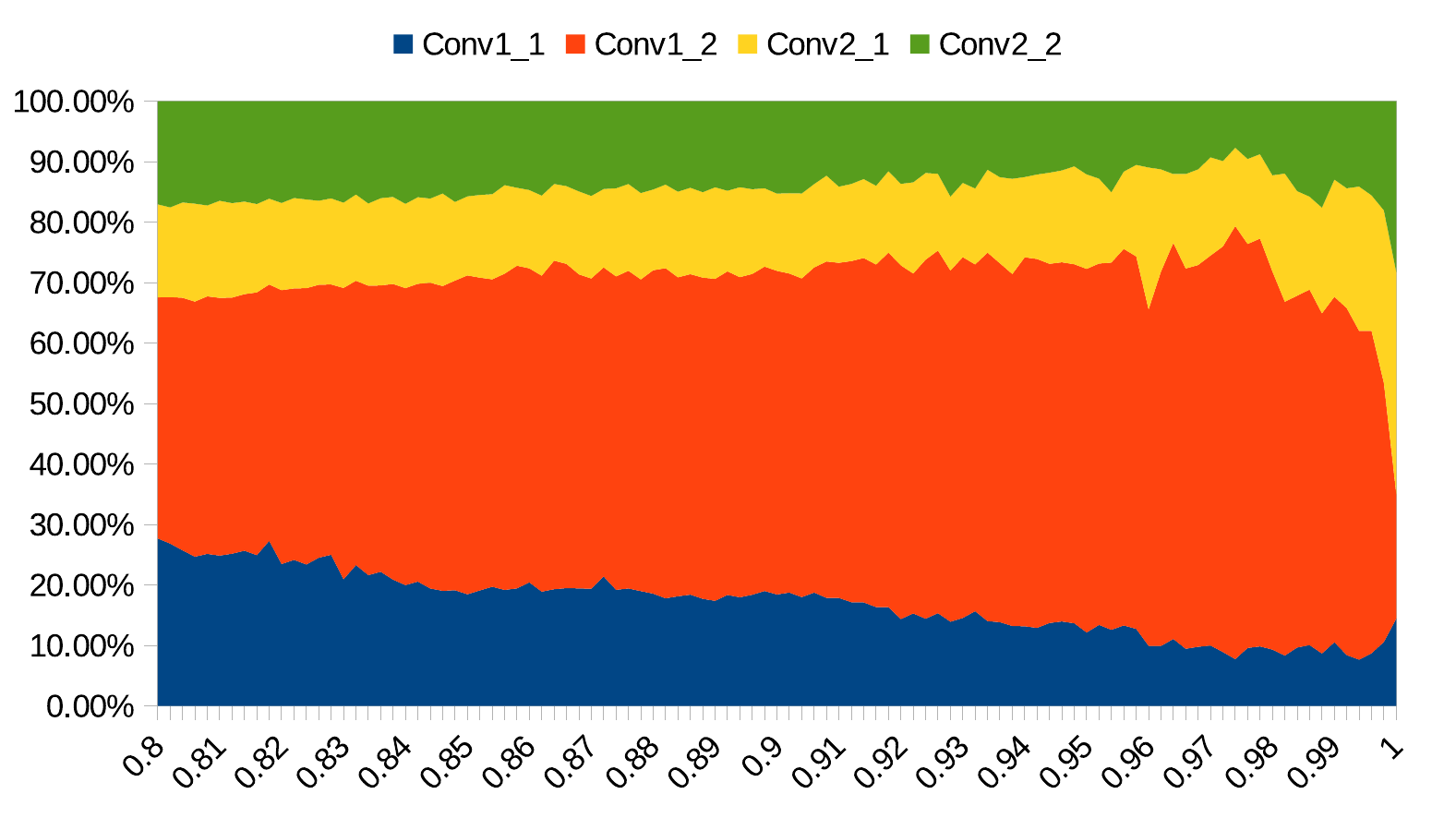}
\caption{energy distribution comparison of layer ``conv1\_1'', ``conv1\_2'', ``conv2\_1'' and ``conv2\_2''. The horizontal axis represents normalized magnitude from 0.8 to 1, and the area shows the comparison of each layer's normalized energy.}
\label{fig:energy_count}
\end{figure}

We calculated the energy distribution of layer ``conv1\_2'' as it induces the largest deviation, layer ``conv1\_1'', ``conv2\_1'' and ``conv2\_2'' are also tested as reference. Figure \ref{fig:energy_count} reveals that, compared with other two layers, the energy of layer ``conv1\_2'' is more concentrated at large value, which indicates stronger correlated.
\section {Conclusion}
\noindent In this paper, we designed a CNN accelerator that substituted floating point representation with BFP representation. Using BFP, the burdensome floating-point arithmetics in convolution layers, which is the majority of the overall CNN architecture, are replaced by the light fixed-point arithmetics. Using 8-bit mantissa, the worst accuracy drop of deep neural networks is less than 0.3\% without retraining. In addition, we developed the NSR upper bound analytical model with the largest deviation less than 8.9dB, which provides the guidance for hardware design. 

\section{Acknowledgement}
This work is supported by the National Key Research and Development (2016YFB0200505).
{\small
\fontsize{9.5pt}{10.5pt} \selectfont
\bibliographystyle{aaai}\bibliography{reference}
}
\end{document}